\documentclass[letterpaper]{article} 
\usepackage{aaai25}  
\usepackage{times}  
\usepackage{helvet}  
\usepackage{courier}  
\usepackage[hyphens]{url}  
\usepackage{graphicx} 
\usepackage{natbib}  
\usepackage{caption} 
\usepackage{algorithm}
\usepackage{algorithmic}

\usepackage{newfloat}
\usepackage{listings}

\usepackage{bm}
\usepackage{amsfonts}
\usepackage{amsmath}
\usepackage{multirow}
\usepackage{rotating}
\usepackage{diagbox}
\usepackage{makecell}
\usepackage{colortbl}

\usepackage{enumitem}
\usepackage{ifthen}
\usepackage{amssymb}
\newlist{checklist}{itemize}{1}
\newcommand{\checkbox}[1]{\ifthenelse{\equal{#1}{true}}{$\boxtimes$}{$\square$}}
\setlist[checklist]{label=\checkbox{false}}

\usepackage{color}

\newcommand{\ie}{i.e.,~}

\newcommand{\red}[1]{\textcolor{red}{#1}}

\newcommand{\addref}[1]{\red{[==]}}

\newlist{mymargins}{itemize}{1}
\setlist[mymargins]{leftmargin=2em, label={}, itemindent=0em}

\DeclareCaptionStyle{ruled}{labelfont=normalfont,labelsep=colon,strut=off} 
\floatstyle{ruled}
\newfloat{listing}{tb}{lst}{}
\floatname{listing}{Listing}
\title{Rethinking Graph Transformer Architecture Design for Node Classification}
\author{
    Jiajun Zhou\textsuperscript{\rm 1,2,3},
    Xuanze Chen\textsuperscript{\rm 1,2,3},
    Chenxuan Xie\textsuperscript{\rm 1,2},
    Shanqing Yu\textsuperscript{\rm 1,2},
    Qi Xuan\textsuperscript{\rm 1,2},
    Xiaoniu Yang\textsuperscript{\rm 1,4}
}
\affiliations{
    \textsuperscript{\rm 1}Institute of Cyberspace Security, Zhejiang University of Technology, Hangzhou 310023, China\\
    \textsuperscript{\rm 2}Binjiang Institute of Artificial Intelligence, ZJUT, Hangzhou 310056, China\\
    \textsuperscript{\rm 3}College of Computer Science and Technology, Zhejiang University of Technology,  Hangzhou 310023, China\\
    \textsuperscript{\rm 4}Science and Technology on Communication Information Security Control Laboratory, Jiaxing 314033, China

}

\usepackage{bibentry}

\begin{document}

\maketitle

\begin{abstract}

Graph Transformer (GT), as a special type of Graph Neural Networks (GNNs), utilizes multi-head attention to facilitate high-order message passing. However, this also imposes several limitations in node classification applications: 1) nodes are susceptible to global noise; 2) self-attention computation cannot scale well to large graphs. In this work, we conduct extensive observational experiments to explore the adaptability of the GT architecture in node classification tasks and draw several conclusions: the current multi-head self-attention module in GT can be completely replaceable, while the feed-forward neural network module proves to be valuable. Based on this, we decouple the propagation ($\mathbf{P}$) and transformation ($\mathbf{T}$) of GNNs and explore a powerful GT architecture, named GNNFormer, which is based on the P/T combination message passing and adapted for node classification in both homophilous and heterophilous scenarios. Extensive experiments on 12 benchmark datasets demonstrate that our proposed GT architecture can effectively adapt to node classification tasks without being affected by global noise and computational efficiency limitations.

\end{abstract}

%

\section{Introduction}

Transformer can alleviate certain limitations encountered by Graph Neural Networks (GNNs) in graph-related tasks, such as over-squashing~\cite{alon2020bottleneck}, long-range dependency~\cite{graphgps, exphormer, grit}, weak connectivity~\cite{Nodeformer, Graphormer}, and limited expressiveness~\cite{san, graphit, grover}. These advantages have motivated extensive research on Graph Transformer (GT), demonstrating its strong potential in tasks such as graph classification, link prediction, and node classification.

Despite the superior performance of GT over classical GNNs in certain scenarios, its adaptation for the prevalent node classification task exhibits several limitations, which can be summarized as follows:
\textbf{1) Self-attention mechanism inevitably introduces noise.} Previous studies ~\cite{pathmlp, h2gcn, mixhop} suggest that shallow GNNs can only capture local neighborhood information of nodes and cannot effectively access higher-order information. Instead, GTs employ the self-attention mechanism to receive information from all other nodes, thereby capturing higher-order relationships such as long-range dependencies. However, self-attention actually disregards the topological structure of the graph and treats it as fully connected~\cite{Graphormer}, which inevitably introduces irrelevant information, or noise, especially in homophilous scenarios. 
Thus, the adaptability of GTs relying on self-attention in node classification task is still open to question.
\textbf{2) The self-attention mechanism is unsuitable for large-scale graph computation.} For vanilla GT, the computational complexity of self-attention is quadratic with respect to the number of nodes. In real-world scenarios such as social networks and e-commerce recommendations, graph-structured data often contains a massive number of nodes, making the direct application of the self-attention mechanism computationally and memory-intensive. Therefore, GTs relying on self-attention are highly inefficient for node classification on large-scale graphs.

For the former limitation, several studies~\cite{min2022transformer,dwivedi2020generalization} utilize topological information as masks or biases to recalibrate the self-attention distribution, thereby reducing the introduction of noise to some extent. Regarding the latter limitation, several studies~\cite{Nodeformer,SGFormer,ASN-GT,Gapformer} attempt to improve the efficiency of GTs on large-scale graphs by reducing the computational complexity of attention computation or compressing the graph size. 
\emph{Despite the observed performance improvement, these studies persist in applying the GT architecture to node classification tasks without thoroughly investigating whether the current GT architecture is genuinely adaptability for this purpose.}

\begin{figure}[!htb]
  \centering
  \includegraphics[width=\linewidth]{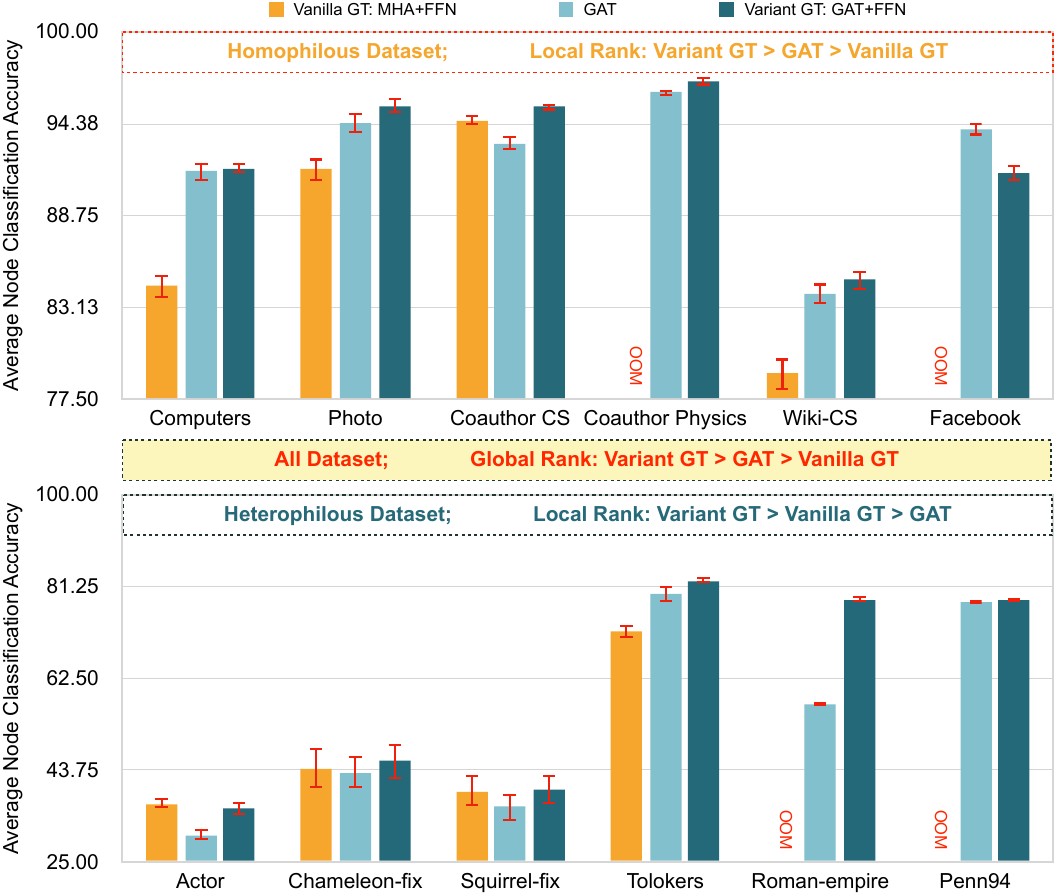}
  \caption{
  Comparison of node classification performance between vanilla GT, vanilla GNN, and variant GT.}
  \label{fig: obs1}
\end{figure}

In this work, we question this adaptability and obtain the supporting observations, as shown in Figure~\ref{fig: obs1}, by substituting the multi-head self-attention module (MHA) in vanilla GT with single graph attention layer (more details will be discussed later).
Specifically, we find that: 1) the applicability of the self-attention module in GT architecture to node classification is not guaranteed; 2) the feed-forward network module (FFN) in GT architecture appears to be beneficial for node classification.
%
Based on extensive observations, we further disentangle message passing into propagation $\mathbf{P}$ and transformation $\mathbf{T}$, and design a graph transformer architecture called GNNFormer that allows for flexible combination of $\mathbf{P}$ and $\mathbf{T}$, making it suitable for node classification in both homophilous and heterophilous scenarios. 
The main differences compared to the vanilla GT architecture are as follows: 1) substituting the multi-head self-attention module with stackable message-passing blocks (PT-block); 2) optimizing the feed-forward network with SwishGLU; 3) improving the residual connections to adaptive initial residual connections. 
The incorporation of three key design in our GNNFormer enables the mitigation of global noise interference during message passing, facilitates effective fusion of graph attribute and structural features, and ultimately achieves powerful node representation learning.
Extensive experiments on 12 homophilous and heterophilous benchmark datasets validate the effectiveness and superiority of our GNNFormer framework in node classification tasks, compared to existing graph transformers.



\section{Preliminaries and Related Work}
A graph is denoted as $G = (V, E, \boldsymbol{X}, \boldsymbol{Y})$, where $V$ and $E$ are the set of nodes and edges respectively, $\boldsymbol{X} \in \mathbb{R}^{|V| \times d}$ is the node feature matrix, and $\boldsymbol{Y} \in\mathbb{R}^{|V| \times C}$ is the node label matrix. Each node is associated with a feature vector $\boldsymbol{x} \in \mathbb{R}^d$ and a one-hot encoded label vector $\boldsymbol{y}\in\mathbb{R}^C$. Here we use $|V|$, $d$ and $C$ to denote the number of nodes, the dimension of the node features, and the number of classes, respectively. The graph topology information $(V, E)$ can also be represented by an adjacency matrix $\boldsymbol{A} \in \mathbb{R}^{|V| \times |V|}$, where $\boldsymbol{A}_{ij}=1$ indicates the existence of an edge between $v_i$ and $v_j$, and $\boldsymbol{A}_{ij}=0$ otherwise. Based on the adjacency matrix, we can define the degree distribution of $G$ as a diagonal degree matrix $\boldsymbol{D} \in \mathbb{R}^{|V| \times |V|}$ with $\boldsymbol{D}_{ii}=\sum_{j=1}^{|V|}\boldsymbol{A}_{ij}$ representing the degree value of $v_i$.
Node classification is a fundamental task in graph machine learning. It involves assigning labels to the nodes of a graph based on their features and the graph topology structure.

\subsection{Disentanglement for GNN Architecture}\label{sec: pt}

Currently, most existing GNNs follow a unified message-passing framework~\cite{MPNN}, in which the message passing phase is decomposed into three processes: message generation, aggregation, and node feature update.
\begin{equation}
    \begin{aligned}
         \textbf{message:~}& \boldsymbol{m}_{i\leftarrow j}^{(l)}=\text{MES}^{(l)}\left(\boldsymbol{h}_j^{(l-1)}, \boldsymbol{h}_i^{(l-1)}, \boldsymbol{e}_{ji}\right)\\
         \textbf{aggregation:~}& \boldsymbol{m}_i^{(l)}=\text{AGG}\left(\left\{\boldsymbol{m}_{i\leftarrow j}^{(l)}\mid j\in\mathcal{N}(i) \right\}\right)\\
         \textbf{update:~}& \boldsymbol{h}_i^{(l)}=\text{UPD}^{(l)}\left(\boldsymbol{h}_i^{(l-1)}, \boldsymbol{m}_i^{(l)}\right)\\
    \end{aligned}
\end{equation}
where $\boldsymbol{m}_{i\leftarrow j}^{(l)}$ denotes the message sent from node $v_j$ to $v_i$ at iteration step $l$, and depends on the feature $\boldsymbol{h}_j^{(l-1)}$ of the sending node, the feature $\boldsymbol{h}_i^{(l-1)}$ of the receiving node and the feature $\boldsymbol{e}_{ji}$ of edge between them. The message function $\text{MES}^{(l)}$ can be a parameterized model such as MLPs. $\text{AGG}$ is the aggregation function, such as summation, averaging and maximum, which is used to aggregate messages from the neighborhood $\mathcal{N}(i)$ of the target node $v_i$. The update function $\text{UPD}^{(l)}$ can be a neural network that integrates the current node state and the aggregated messages to produce a new node state. 
From a more decoupled perspective, the message-passing phase can be decomposed into two functionally independent operations: Propagation and Transformation~\cite{PT}.
\begin{equation}
    \begin{aligned}
         &\textbf{prop:~} \boldsymbol{h}_i^{(l)} = \mathbf{P}\left(\boldsymbol{h}_i^{(l-1)}, \left\{\boldsymbol{h}_{j}^{(l-1)}, \boldsymbol{e}_{ji}\mid j\in\mathcal{N}(i) \right\}\right)\\
         &\textbf{trans:~} \boldsymbol{h}_i^{(l)}=\mathbf{T}\left(\boldsymbol{h}_i^{(l)}\right)\\
    \end{aligned}
\end{equation}
where $\mathbf{P}$ is the propagation function that combines message generation and aggregation from neighbor node $v_j$ to target node $v_i$. $\mathbf{T}$ performs a non-linear transformation on the state of the nodes after propagation.
Based on the disentanglement, existing GNN architectures can be roughly and loosely categoried into four types according to the stacking order of propagation and transformation operations: $\mathbf{PTPT}$, $\mathbf{PPTT}$, $\mathbf{TTPP}$, and $\mathbf{TPTP}$, as listed in Table~\ref{tab: PT}.

\begin{table}
    \centering
    \caption{Disentanglement for existing model architectures.}
    \label{tab: PT}
    \renewcommand\arraystretch{1.2}
    \resizebox{\linewidth}{!}{
    \begin{tabular}{c|c|c|c} 
    \hline
    \multicolumn{2}{c|}{Catrgory} & Method                & Message Passing  \\ 
    \hline
    NN                   & $\mathbf{TT}$     & MLP, LINKX              & $\mathbf{T}(\mathbf{T}(\boldsymbol{X}))$          \\ 
    \hline
    \multirow{4}{*}{GNN} & $\mathbf{PTPT}$   & GCN, GAT, SAGE, FSGNN   & $\mathbf{T}(\mathbf{P}(\mathbf{T}(\mathbf{P}(\boldsymbol{X}))))$    \\ 
    \cline{2-4}
                         & $\mathbf{PPTT}$   & ACMGCN                  & $\mathbf{T}(\mathbf{T}(\mathbf{P}(\mathbf{P}(\boldsymbol{X}))))$    \\ 
    \cline{2-4}
                         & $\mathbf{TTPP}$   & GPRGNN, H2GCN, GloGNN   & $\mathbf{P}(\mathbf{P}(\mathbf{T}(\mathbf{T}(\boldsymbol{X}))))$    \\ 
    \cline{2-4}
                         & $\mathbf{TPTP}$   & FAGCN                   & $\mathbf{P}(\mathbf{T}(\mathbf{P}(\mathbf{T}(\boldsymbol{X}))))$    \\
    \hline
    \end{tabular}}
\end{table}

\subsection{Graph Transformer for Node Classification}

The graph transformer, serving as a new graph representation learning architecture, is proposed to alleviate the limitations of GNNs in dealing with issues such as over-squashing~\cite{alon2020bottleneck}, long-range dependency~\cite{graphgps, exphormer}, weak connectivity~\cite{Nodeformer, Graphormer}, etc. Generally, the vanilla GT architecture broadly follows the classic Transformer~\cite{Transformer} and consists of two essential modules: a multi-head self-attention module (MHA) and a feed-forward network (FFN), as shown in Figure~\ref{fig: architecture}(a). The MHA module first transforms the node features into query vectors ($\boldsymbol{Q}$), key vectors ($\boldsymbol{K}$) and value vectors ($\boldsymbol{V}$), then performs the inner-product operation on the former two to compute the attention scores, and ultimately employs the obtained scores for weighted aggregation of the value vector:
\begin{equation}
    \begin{gathered}
    \boldsymbol{Q}=\boldsymbol{H} \boldsymbol{W}_Q,\quad  
    \boldsymbol{K}=\boldsymbol{H} \boldsymbol{W}_K,\quad 
    \boldsymbol{V}=\boldsymbol{H} \boldsymbol{W}_V \\
    \boldsymbol{H}^{\prime}=\operatorname{softmax}\left(\frac{\boldsymbol{Q} \boldsymbol{K}^{\top}}{\sqrt{d^{\prime}}}\right) \boldsymbol{V}
    \end{gathered}
\end{equation}
where $\boldsymbol{H}=\left[\boldsymbol{h}_1^\top,\cdots,\boldsymbol{h}_{|V|}^\top\right]\in\mathbb{R}^{|V|\times d}$ denotes the input node embedding matrix, $\boldsymbol{W}_Q$, $\boldsymbol{W}_K$, $\boldsymbol{W}_V\in\mathbb{R}^{d\times d^{\prime}}$ are the projection matrics, $d^\prime$ is the output hidden dimension.
The FFN follows the MHA to enhance the expressiveness of GT through linear and non-linear transformations. Additionally, residual connection~\cite{res} is employed to prevent the gradient vanishing as the model depth increases, followed by layer normalization~\cite{layernorm} to stabilize model training and accelerate convergence.


In this paper, we primarily focus on research related to GT in node classification.
Graph-BERT~\cite{Graph-Bert} utilizes top-k affinity sampling to extract unlinked subgraphs as training samples, and incorporates three positional embeddings to pre-train a vanilla GT for node representation learning.
Considering the importance of topology structure in graph representation learning, several studies~\cite{dwivedi2020generalization,EGT} integrate edge embeddings as additional inputs to the MHA to recalibrate the attention computation, thereby better utilizing the graph connectivity inductive bias.
Considering the efficiency limitations of vanilla GT on large-scale graphs, researchers have attempted to enhance the scalability of GT by reducing the computational complexity of attention or compressing the graph size. NodeFormer~\cite{Nodeformer} introduces a kernelized Gumbel-Softmax operator to reduce the quadratic complexity of self-attention computation to linear complexity. SGFormer~\cite{SGFormer} designs a single-layer, single-head global attention model, avoiding the stacking of multiple multi-head attention layers, significantly reducing computational complexity. NAGphormer~\cite{NAGphormer} transforms the 1- to k-hop neighborhood representation of each node into a sequence of tokens and trains it in a mini-batch manner to improve scalability to large-scale graphs. ASN-GT~\cite{ASN-GT} and GapFormer~\cite{Gapformer} first utilize graph coarsening or pooling techniques to significantly compress the number of nodes in large-scale graphs, then train GT on the coarsened graph, thereby reducing computational overhead.

\begin{figure}[!htb]
    \centering
    \includegraphics[width=\linewidth]{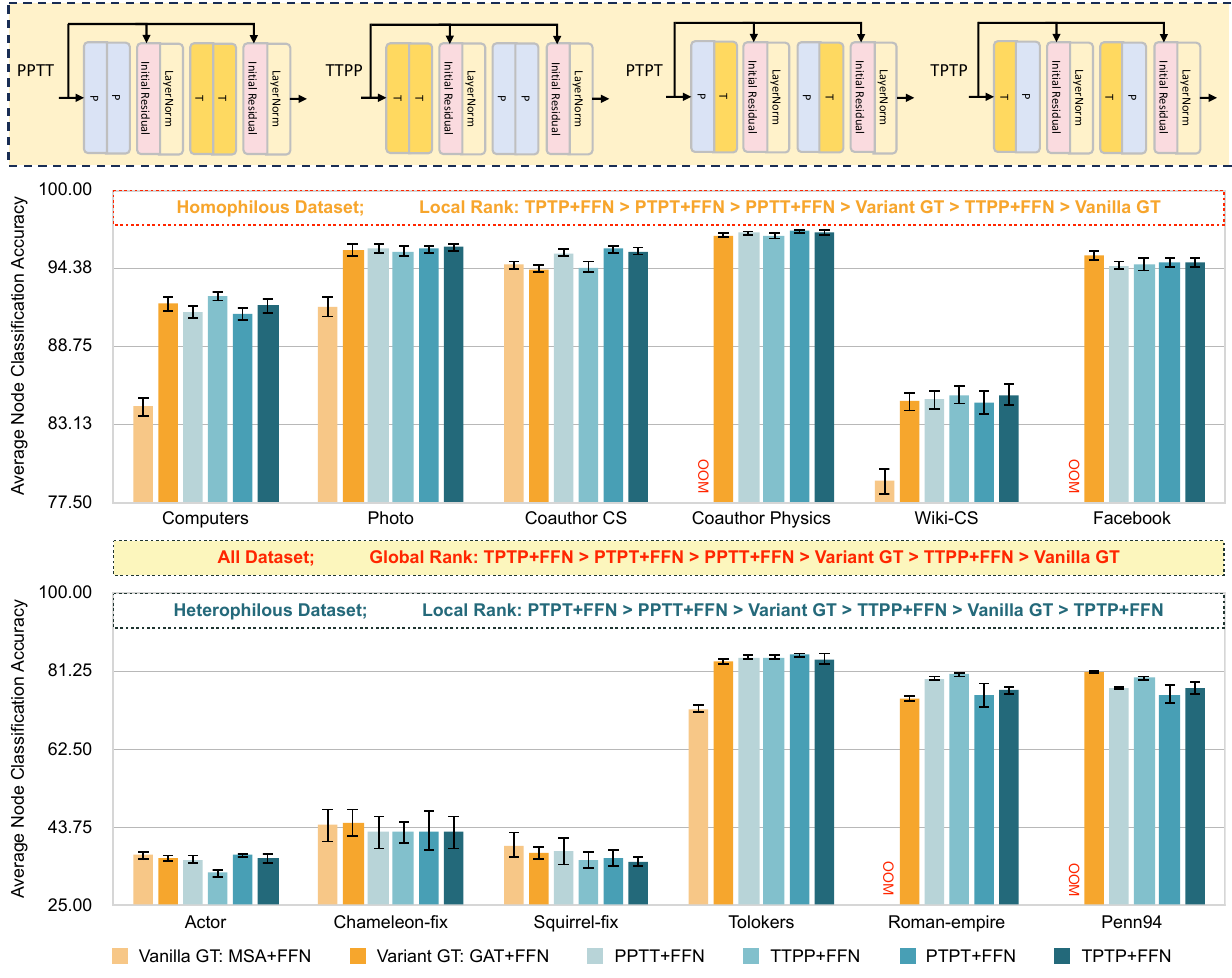}
    \caption{
    Comparison of node classification performance between different message passing architectures.}
    \label{fig: obs2}
\end{figure}
\section{Motivating Observations}\label{sec: obs}
\subsection{Vanilla GT for Node Classification Is Debatable}
Compared to GNNs, GT actually employs self-attention to achieve global message passing.
However, this practice also reveals the limitations of GT in node classification, which can be summarized as follows:

\noindent\textbf{Proposition 1.~} \emph{The self-attention mechanism in the vanilla GT architecture actually treats the graph as fully connected for global message propagation, inevitably introducing noise from irrelevant nodes. Additionally, the quadratic computational complexity of the attention mechanism makes GT inefficient for node classification on large-scale graphs. Therefore, the adaptability of the vanilla GT architecture for node classification tasks remains debatable.}

\noindent\textbf{Observation 1.~}
To illustrate our proposition, we first train a vanilla GT and a two-layer GAT~\cite{GAT}, using common settings of 4 attention heads, 64 hidden dimensions, 0.5 dropout rate, ReLU activation and 10 runs. 
The performance comparison of node classification on homophilous and heterophilous datasets is shown in Figure~\ref{fig: obs1}. It is evident that vanilla GT underperforms GAT on 8 out of the 12 benchmarks. Further, by replacing the MHA in vanilla GT with a single graph attention layer, we derive the variant GT that demonstrates not only greater stability but also significant performance improvements compared to vanilla GT in most cases. This indicates that the self-attention mechanism indeed introduces irrelevant noise during message passing, thereby limits model performance. 
Additionally, the variant GT outperforms GAT in most cases, suggesting that incorporating a subsequent FFN offers valuable guidance for designing GT architectures more suited for node classification, meriting further exploration.
Finally, vanilla GT also experiences out of memory (OOM) issue on several datasets, further highlighting its efficiency limitations.



\begin{figure*}[!htb]
    \centering
    \includegraphics[width=0.85\textwidth]{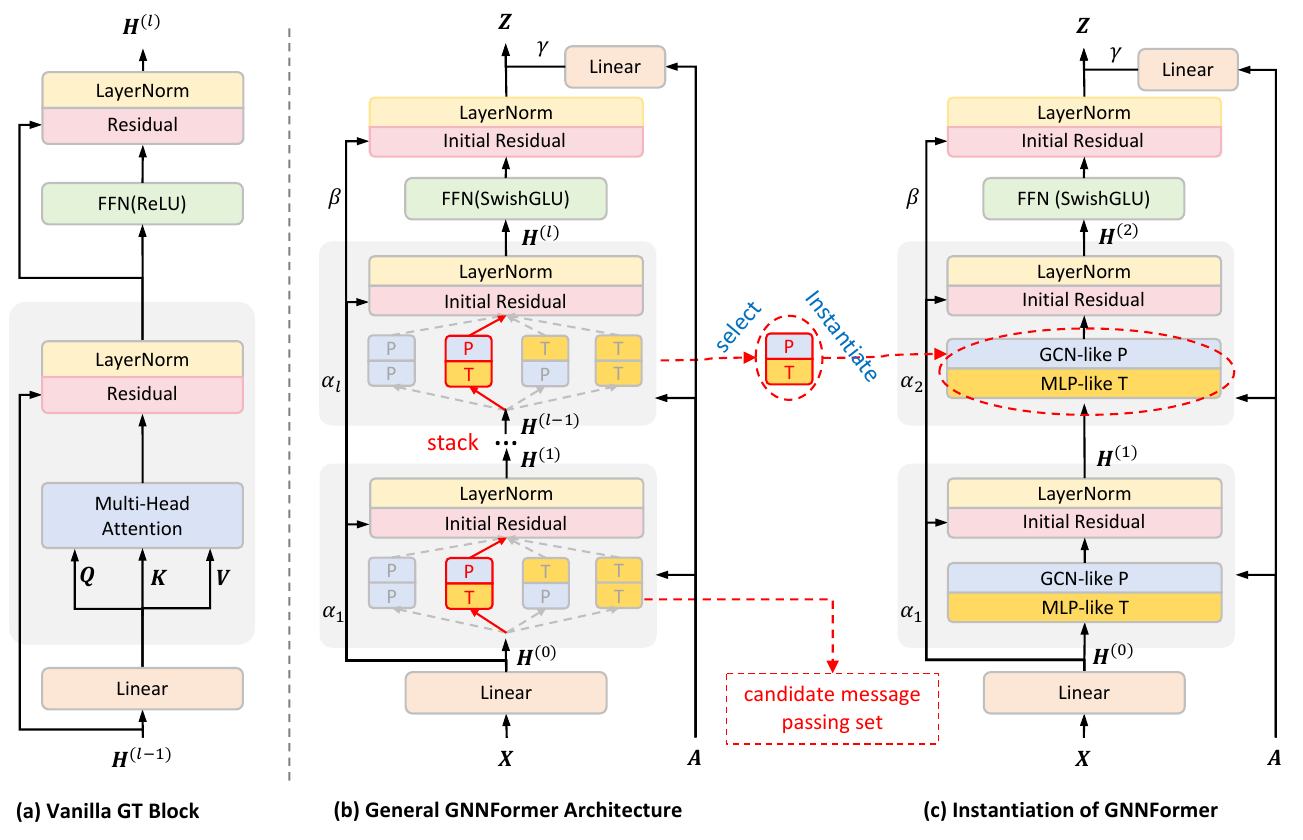}
    \caption{Illustration of graph transformer architectures.}
    \label{fig: architecture}
\end{figure*}

\subsection{Exploring Message Passing in GT Is Valuable}

In this work, we regard GT and its variants as another form of GNNs. Based on the disentangled GNN architectures mentioned above, we can further approximate the vanilla GT architecture in Figure~\ref{fig: architecture}(a) as a $\mathbf{TPTT}$ form, where the Linear, MHA, and FFN modules can be respectively approximated as $\mathbf{T}$, $\mathbf{P}$, and $\mathbf{TT}$. Considering that the instantiation of $\mathbf{P}$ with multi-head self-attention leads to unstable performance of GT in node classification, this drives us to rethink the design of the GT architecture.

\noindent\textbf{Proposition 2.~} 
\emph{To enhance the adaptability of the GT architecture for node classification tasks, rethinking the message passing design to replace the original MHA module is a strightforward practice.}

\noindent\textbf{Observation 2.~} 
In Observation 1, we directly substitute the MHA module in vanilla GT with a single multi-head graph attention layer, yielding a variant GT. This crude architectural assembly is naive, although it achieves commendable performance. Based on this, we further explore message passing designs based on the P/T combination (PT-block) that can replace the MHA in vanilla GT. We substitute the MHA in vanilla GT with different PT-blocks ($\mathbf{PTPT}$, $\mathbf{PPTT}$, $\mathbf{TTPP}$, and $\mathbf{TPTP}$, with P instantiated by GAT-like propagation). These blocks use initial residual connections and layer normalization, making the resulting variants more aligned with the Transformer architecture. As shown in Figure~\ref{fig: obs2}, all models share common settings: 4 attention heads, 64 hidden dimension, 0.5 dropout rate, ReLU activation and 10 runs.
We can observe that these P/T-based variants outperform the naively assembled variant GT (GAT+FFN) in most cases, indicating that: 1) Message passing modules based on the P/T combination can be more flexibly integrated into the GT architecture; 2) Certain architectural components of Transformers, such as residual connections, also contribute to node representation learning.
Moreover, the varying performance of different P/T-blocks across different types of datasets further motivates us to explore more universal message passing designs within GT architectures.

\section{GNNFormer: New GT Architecture Design}


Based on the aforementioned motivating observations, we propose a novel graph transformer architecture named GNNFormer, specifically tailored for node classification. The architecture is illustrated in Figure~\ref{fig: architecture}(b). Compared to the vanilla GT block, the modifications are mainly reflected in three aspects: 1) the original multi-head self-attention module is substituted with the stackable P-T combination message passing block (PT-block); 2) the FFN is optimized via SwishGLU; 3) the original residual connection is improved to the initial residual connection. The design details of GNNFormer will be discussed below.

As shown in Figure~\ref{fig: architecture}(a), vanilla GT learns node representations by stacking transformer blocks, with the input to the current block being the output of the previous block. Differently, our GNNFormer consists of only a single transformer block, which takes node features and adjacency information as input and outputs the final node representations.

First, the input features $\boldsymbol{X}$ will be transformed into an initial feature embedding through a linear transformation parameterized by $\boldsymbol{W}_0\in\mathbb{R}^{d\times d^\prime}$ and a ReLU activation:
\begin{equation}
\boldsymbol{H}^{(0)}=\operatorname{ReLU}\left(\boldsymbol{X}\boldsymbol{W}_0\right)
\end{equation}
where $d^\prime$ is the hidden dimension.
Next, we stack message passing blocks, \ie PT-blocks, to further learn node representations. Each PT-block consists of a message passing function $f$, an initial residual connection, and a layer normalization operation $\operatorname{LN}\left( \cdot \right)$. For the $l$-th PT-block, it takes the adjacency matrix $\boldsymbol{A}$ and the node representations $\boldsymbol{H}^{(l-1)}$ output by the previous PT-block as input, and outputs new node representations via propagation and transformation:
\begin{equation}
   \boldsymbol{H}^{(l)}=\operatorname{LN}\left(\alpha_l \cdot \boldsymbol{H}^{(0)}+(1-\alpha_l)\cdot f\left(\boldsymbol{H}^{(l-1)}, \boldsymbol{A}\right)\right)\\
\end{equation}
where $f$ is selected from the candidate message passing set \{$\mathbf{PP}$, $\mathbf{PT}$, $\mathbf{TP}$, $\mathbf{TT}$\}, and $\alpha_l$ is a learnable parameter that controls the adaptive residual connection.



After message passing through $l$ PT-blocks, GNNFormer has effectively fused the attribute information of the nodes with the topological information. Furthermore, motivated by observation 1, which suggests that appending an FFN can enhance the expressive power of vanilla GNNs, we preserve the FFN module in the GNNFormer architecture. Specifically, the node representations $\boldsymbol{H}^{(l)}$ will first be fed into a two-layer FFN that employs the SwishGLU in place of the first linear layer and the activation function:
\begin{equation}
    \boldsymbol{Z}^\prime=\left( \operatorname{Swish}\left(\boldsymbol{H}^{(l)}\boldsymbol{W}_1\right)\otimes \boldsymbol{H}^{(l)}\boldsymbol{W}_2 \right) \boldsymbol{W}_3
\end{equation}
where $\boldsymbol{W}_1$, $\boldsymbol{W}_2$, $\boldsymbol{W}_3\in\mathbb{R}^{d^\prime\times d^\prime}$ are the transformation weight matrices, and $\otimes$ is the element-wise multiplication. 
The SwishGLU~\cite{SwishGLU} combines the smoothness of the Swish~\cite{Swish} activation function with the selectivity of Gated Linear Units~\cite{GLU}. The former ensures more stable gradient updates during back-propagation, while the latter allows the GNNFormer to selectively propagate more important features, especially in heterophilous scenarios. Consequently, the enhanced FFN improves the non-linear expressive power of GNNFormer, enabling it to better capture complex node relationships and feature interactions within various graphs.

Subsequently, an initial residual connection and layer normalization will be applied to the output of the FFN:
\begin{equation}
   \boldsymbol{Z}^\prime=\operatorname{LN}\left(\beta \cdot \boldsymbol{H}^{(0)}+(1-\beta)\cdot \boldsymbol{Z}^\prime\right)\\
\end{equation}
where $\beta$ is a learnable parameter that controls the adaptive residual connection. 
Finally, we further integrate the topology information to obtain the final node representations:
\begin{equation}
    \boldsymbol{Z}=\gamma \cdot \boldsymbol{Z}^\prime+(1-\gamma)\cdot\boldsymbol{A}\boldsymbol{W}_4
\end{equation}
where $\boldsymbol{W}_4\in\mathbb{R}^{|V| \times d^\prime}$ is the weight matrix that encodes the topology information, and $\gamma$ is a learnable parameter that controls the contribution of topology information. 

To achieve node classification, we finally use a prediction head $f_\text{pred}$ parameterized by $\boldsymbol{W}_5\in\mathbb{R}^{d^\prime \times C}$ and Softmax activation to obtain the node predictions:
\begin{equation}
    \hat{\boldsymbol{Y}}=\operatorname{Softmax}\left( \boldsymbol{Z}\boldsymbol{W}_5 \right)
\end{equation}
During model training, binary cross-entropy classification loss is used as the optimization objective:
\begin{equation}
    \mathcal{L} = -\operatorname{trace}\left(\boldsymbol{Y}_\text{train}^\top \cdot \log \hat{\boldsymbol{Y}}_\text{train} \right)
\end{equation}
where the trace operation $\operatorname{trace}\left( \cdot \right)$ is used to compute the sum of the diagonal elements of the matrix.

\section{Experiments}
\subsection{Datasets}
We conduct extensive exploratory and evaluation experiments on 12 benchmark datasets, which include (1) Six homophilous datasets: Computers, Photo~\cite{mcauley2015I}, Coauthor CS, Coauthor Physics~\cite{Shchur2018PitfallsOG}, Wiki-CS~\cite{mernyei2020wiki}, and Facebook~\cite{rozemberczki20201R};
(2) Six heterophilous datasets: Actor~\cite{tang2009S}, Chameleon-fix, Squirrel-fix, Tolokers, Roman-empire~\cite{platonov2023a}, and Penn94~\cite{NEURIPS2021_ae816a80}.
All datasets are divided into training, validation, and testing sets in a proportion of 48\%: 32\%: 20\%. More details are presented in Appendix A.


\subsection{Baselines}
To evaluate the effectiveness and superiority of the new GT architecture, we compare it against 15 baselines, including (1) MLP; (2) Classical GNNs: GCN~\cite{GCN}, GAT~\cite{GAT}, GraphSAGE~\cite{GraphSAGE}; (3) Heterophilous GNNs: LINKX~\cite{NEURIPS2021_ae816a80}, H2GCN~\cite{H2GCN2020}, GPRGNN~\cite{chien2021adaptive}, FAGCN~\cite{fagcn2021}, ACMGCN~\cite{luan2022revisiting}, GloGNN~\cite{li2022finding}, FSGNN~\cite{MAURYA2022101695}; (4) Graph Transformers: vanilla GT, ANS-GT~\cite{ASN-GT}, SGformer~\cite{SGFormer}, NAGphormer~\cite{NAGphormer}.
More details are presented in Appendix B.

\subsection{Experimental Settings}
To ensure the reproducibility of our experiments, we utilize 10 random seeds to fix the data splits and model initialization, and report the average accuracy and standard deviation over 10 experimental runs.
For all methods, we set the search space of common parameters as follows:
maximum epochs to 500 with 100 patience,
hidden dimension $d^\prime$ to 64,
optimizer to AdamW~\cite{AdamW},
learning rate in \{0.005, 0.01, 0.05, 0.1\},
dropout rate in \{0.1, 0.3, 0.5, 0.7, 0.9\}.
For our GNNFormer, we search the number of PT-blocks in \{1, 2, 3\}. 
For all baselines, we search the common parameters in the same parameter spaces. 
And more default parameter details of baselines are reported in Appendix C.
Moreover, GNNFormer are implemented in PyTorch 1.11.0, Pytorch-Geometric 2.1.0 with CUDA 12.0 and Python 3.9.
All experiments are conducted at NVIDIA A100 40GB.


\begin{table*}
    \centering
    \caption{Node classification results: average test accuracy (\%) $\pm$ standard deviation. The best results are highlighted in bold, while the second-best results are underlined. ``Local Rank'' indicates the average performance ranking across homophilous or heterophilous datasets, ``Global Rank'' indicates the average performance ranking across all datasets.}
    \label{tab: main}
    \renewcommand\arraystretch{1.2}
    \resizebox{\textwidth}{!}{
        \begin{tabular}{c|c|ccccccc|ccccccc|c}
            \hline
            \multicolumn{2}{l|}{\diagbox{Method}{Dataset}}                                          & Computers             & Photo                 & \begin{tabular}[c]{@{}c@{}}Coauthor\\CS\end{tabular} & \begin{tabular}[c]{@{}c@{}}Coauthor\\Physics\end{tabular} & Wiki-CS               & Facebook              & \begin{tabular}[c]{@{}c@{}}Local\\Rank\end{tabular} & Actor                 & \begin{tabular}[c]{@{}c@{}}Chameleon\\-fix\end{tabular} & Squirrel-fix          & Tolokers              & \begin{tabular}[c]{@{}c@{}}Roman\\-empire\end{tabular} & Penn94                & \begin{tabular}[c]{@{}c@{}}Local\\Rank\end{tabular} & \multicolumn{1}{c}{\begin{tabular}[c]{@{}c@{}}Global\\Rank\end{tabular}}  \\ 
            \hline
            \multirow{4}{*}{Vanilla}                                                  & MLP         & 85.01 $\pm$ 0.84          & 92.00 $\pm$ 0.56          & 94.80 $\pm$ 0.35                                         & 96.11 $\pm$ 0.14                                              & 79.57 $\pm$ 0.81          & 76.86 $\pm$ 0.34          & 25.17                                              & 37.14 $\pm$ 1.06          & 33.31 $\pm$ 2.32                                            & 34.47 $\pm$ 3.09          & 53.18 $\pm$ 6.35          & 65.98 $\pm$ 0.43                                           & 75.18 $\pm$ 0.35          & 23.83                                              & 24.50                                                                      \\
                                                                                      & GCN         & 91.17 $\pm$ 0.54          & 94.26 $\pm$ 0.59          & 93.40 $\pm$ 0.45                                         & 96.37 $\pm$ 0.20                                              & 83.80 $\pm$ 0.66          & 93.98 $\pm$ 0.34          & 21.50                                              & 30.65 $\pm$ 1.06          & 41.85 $\pm$ 3.22                                            & 33.89 $\pm$ 2.61          & 70.34 $\pm$ 1.64          & 50.76 $\pm$ 0.46                                           & 80.45 $\pm$ 0.27          & 24.17                                              & 22.83                                                                      \\
                                                                                      & GAT         & 91.44 $\pm$ 0.43          & 94.42 $\pm$ 0.61          & 93.20 $\pm$ 0.64                                         & 96.28 $\pm$ 0.31                                              & 83.99 $\pm$ 0.73          & 94.03 $\pm$ 0.36          & 20.83                                              & 30.58 $\pm$ 1.18          & 43.31 $\pm$ 3.42                                            & 36.27 $\pm$ 2.12          & 79.93 $\pm$ 0.77          & 57.34 $\pm$ 1.81                                           & 78.10 $\pm$ 1.28          & 23.00                                              & 21.92                                                                      \\
                                                                                      & GraphSAGE   & 90.94 $\pm$ 0.56          & 95.41 $\pm$ 0.45          & 94.17 $\pm$ 0.46                                         & 96.69 $\pm$ 0.23                                              & 84.75 $\pm$ 0.64          & 93.72 $\pm$ 0.35          & 19.17                                              & 37.60 $\pm$ 0.95          & 44.94 $\pm$ 3.67                                            & 36.61 $\pm$ 3.06          & 82.37 $\pm$ 0.64          & 77.77 $\pm$ 0.49                                           & OOM                       & 16.83                                              & 18.00                                                                     \\ 
            \hline
            \multirow{7}{*}{Heterophilous}                                            & H2GCN       & 91.69 $\pm$ 0.33          & 95.59 $\pm$ 0.48          & 95.62 $\pm$ 0.27                                         & 97.00 $\pm$ 0.16                                              & 84.62 $\pm$ 0.66          & 94.36 $\pm$ 0.32              & 13.33                                              & 37.27 $\pm$ 1.27          & 43.09 $\pm$ 3.85                                            & 40.07 $\pm$ 2.73          & 81.34 $\pm$ 1.16          & 79.47 $\pm$ 0.43                                           & 75.91 $\pm$ 0.44          & 17.33                                               & 15.33                                                                      \\
                                                                                      & GPRGNN      & 91.80 $\pm$ 0.55          & 95.44 $\pm$ 0.33          & 95.17 $\pm$ 0.34                                         & 96.94 $\pm$ 0.20                                              & 84.84 $\pm$ 0.54          & \underline{94.84 $\pm$ 0.24}  & 12.33                                              & 36.89 $\pm$ 0.83          & 44.27 $\pm$ 5.23                                            & 40.58 $\pm$ 2.00          & 73.84 $\pm$ 1.40          & 67.72 $\pm$ 0.63                                           & 84.34 $\pm$ 0.29          & 17.67                                               & 15.00                                                                      \\
                                                                                      & FAGCN       & 89.54 $\pm$ 0.75          & 94.44 $\pm$ 0.62          & 94.93 $\pm$ 0.22                                         & 96.91 $\pm$ 0.27                                              & 84.47 $\pm$ 0.75          & 91.90 $\pm$ 1.95              & 21.00                                             & 37.59 $\pm$ 0.95          & 45.28 $\pm$ 4.33                                            & 41.05 $\pm$ 2.67          & 81.38 $\pm$ 1.34          & 75.83 $\pm$ 0.35                                           & 79.01 $\pm$ 1.09           & 14.00                                               & 17.50                                                                      \\
                                                                                      & ACMGCN      & 91.66 $\pm$ 0.78          & 95.42 $\pm$ 0.39          & 95.47 $\pm$ 0.33                                         & 97.00 $\pm$ 0.27                                              & 85.10 $\pm$ 0.77          & 94.27 $\pm$ 0.33              & 13.83                                             & 36.89 $\pm$ 1.13          & 43.99 $\pm$ 2.02                                            & 36.58 $\pm$ 2.75          & 83.52 $\pm$ 0.87          & 81.57 $\pm$ 0.35                                           & 83.01 $\pm$ 0.46           & 17.67                                               & 15.75                                                                      \\
                                                                                      & GloGNN      & 89.48 $\pm$ 0.63          & 94.34 $\pm$ 0.58          & 95.32 $\pm$ 0.29                                         & OOM                                                       & 80.59 $\pm$ 0.53          & 84.57 $\pm$ 0.62                  & 23.67                                              & 37.30 $\pm$ 1.41          & 41.46 $\pm$ 3.89                                            & 37.66 $\pm$ 2.12          & 58.74 $\pm$ 13.41         & 66.46 $\pm$ 0.41                                           & 85.33 $\pm$ 0.27        & 19.17                                                 & 21.42                                                                      \\
                                                                                      & FSGNN       & 91.03 $\pm$ 0.56          & 95.50 $\pm$ 0.41          & 95.51 $\pm$ 0.32                                         & 96.98 $\pm$ 0.20                                              & 85.10 $\pm$ 0.58          & 94.32 $\pm$ 0.32              & 14.50                                              & 37.14 $\pm$ 1.06          & 45.79 $\pm$ 3.31                                            & 38.25 $\pm$ 2.62          & 83.87 $\pm$ 0.98          & 79.76 $\pm$ 0.41                                           & 83.87 $\pm$ 0.98          & 15.00                                               & 14.75                                                                      \\
                                                                                      & LINKX       & 90.75 $\pm$ 0.36          & 94.58 $\pm$ 0.56          & 95.52 $\pm$ 0.30                                         & 96.93 $\pm$ 0.16                                              & 83.51 $\pm$ 0.78          & 93.84 $\pm$ 0.32              & 18.50                                              & 31.17 $\pm$ 0.61          & 44.94 $\pm$ 3.08                                            & 38.40 $\pm$ 3.54          & 77.55 $\pm$ 0.80          & 61.36 $\pm$ 0.60                                           & 84.97 $\pm$ 0.46          & 19.50                                               & 19.00                                                                      \\ 
            \hline
            \multirow{4}{*}{GT}                                                       & Vanilla GT  & 84.41 $\pm$ 0.72          & 91.58 $\pm$ 0.73          & 94.61 $\pm$ 0.30                                         & OOM                                                       & 79.05 $\pm$ 0.97          & OOM                               & 26.17                                               & 37.08 $\pm$ 1.08          & 44.27 $\pm$ 3.98                                            & 39.55 $\pm$ 3.10          & 72.24 $\pm$ 1.17          & OOM                                                    & OOM                           & 21.17                                               & 23.67                                                                      \\
                                                                                      & ANS-GT      & 90.01 $\pm$ 0.38          & 94.51 $\pm$ 0.24          & 93.93 $\pm$ 0.23                                         & 96.28 $\pm$ 0.19                                              & 83.27 $\pm$ 0.49          & 92.61 $\pm$ 0.16              & 22.67                                               & 37.80 $\pm$ 0.95       & 40.74 $\pm$ 2.26                                            & 36.65 $\pm$ 0.80          & 76.91 $\pm$ 0.85          & 80.36 $\pm$ 0.71                                           & OOM                          & 18.17                                               & 20.42                                                                      \\
                                                                                      & NAGFormer   & 90.22 $\pm$ 0.42          & 94.95 $\pm$ 0.52          & 94.96 $\pm$ 0.25                                         & 96.43 $\pm$ 0.20                                              & 84.31 $\pm$ 0.72          & 93.35 $\pm$ 0.28              & 20.17                                               & 36.99 $\pm$ 1.39          & 46.12 $\pm$ 2.25                                            & 38.31 $\pm$ 2.43          & 66.73 $\pm$ 1.18          & 75.92 $\pm$ 0.69                                           & 73.98 $\pm$ 0.53          & 19.33                                               & 19.75                                                                      \\
                                                                                      & SGFormer    & 90.70 $\pm$ 0.59          & 94.46 $\pm$ 0.49          & 95.21 $\pm$ 0.20                                         & 96.87 $\pm$ 0.18                                              & 82.67 $\pm$ 0.58          & 86.66 $\pm$ 0.54              & 21.17                                               & 36.59 $\pm$ 0.90          & 44.27 $\pm$ 3.68                                            & 38.83 $\pm$ 2.19          & 80.46 $\pm$ 0.91          & 76.41 $\pm$ 0.50                                           & 76.65 $\pm$ 0.49          & 19.00                                               & 20.08                                                                      \\ 
            \hline
            \multirow{3}{*}{\begin{tabular}[c]{@{}c@{}}GNNFormer\\TP+TP\end{tabular}} & GCN-like P  & {\cellcolor[rgb]{0.502,0.502,0.502}}\textbf{92.21 $\pm$ 0.35} & {\cellcolor[rgb]{0.769,0.616,0.18}}\underline{95.90 $\pm$ 0.37}   & {\cellcolor[rgb]{0.655,0.655,0.655}}\underline{95.79 $\pm$ 0.23}  & {\cellcolor[rgb]{0.757,0.592,0.153}}\textbf{97.25 $\pm$ 0.15} & {\cellcolor[rgb]{0.831,0.831,0.831}}84.95 $\pm$ 0.77          & {\cellcolor[rgb]{0.91,0.827,0.573}}94.66 $\pm$ 0.33           & \underline{4.83}                                        & {\cellcolor[rgb]{0.671,0.671,0.671}}37.72 $\pm$ 0.92         & {\cellcolor[rgb]{0.859,0.741,0.408}}\underline{47.13 $\pm$ 2.13}  & {\cellcolor[rgb]{0.957,0.957,0.957}}38.79 $\pm$ 1.72          & {\cellcolor[rgb]{0.761,0.6,0.165}}\underline{85.62 $\pm$ 0.83}    & {\cellcolor[rgb]{0.686,0.686,0.686}}80.80 $\pm$ 0.79          & {\cellcolor[rgb]{0.757,0.592,0.153}}\textbf{85.44 $\pm$ 0.36} & \underline{6.33}                                        & \textbf{5.58}                                                             \\
            & SAGE-like P & {\cellcolor[rgb]{0.616,0.616,0.616}}92.06 $\pm$ 0.47          & {\cellcolor[rgb]{0.769,0.616,0.18}}\underline{95.90 $\pm$ 0.50}   & {\cellcolor[rgb]{0.957,0.957,0.957}}95.64 $\pm$ 0.19          & {\cellcolor[rgb]{0.929,0.847,0.631}}97.13 $\pm$ 0.12          & {\cellcolor[rgb]{0.545,0.545,0.545}}\textbf{85.41 $\pm$ 0.62} & {\cellcolor[rgb]{0.757,0.592,0.153}}\textbf{94.87 $\pm$ 0.18} & \textbf{4.17}                                      & {\cellcolor[rgb]{0.792,0.792,0.792}}37.44 $\pm$ 1.32         & {\cellcolor[rgb]{0.976,0.922,0.776}}46.07 $\pm$ 3.31          & {\cellcolor[rgb]{0.796,0.796,0.796}}40.22 $\pm$ 2.36          & {\cellcolor[rgb]{0.867,0.757,0.439}}84.73 $\pm$ 0.76          & {\cellcolor[rgb]{0.545,0.545,0.545}}\textbf{83.91 $\pm$ 0.55} & {\cellcolor[rgb]{0.773,0.624,0.184}}\underline{85.41 $\pm$ 0.44}         & 7.00                                               & \textbf{5.58}                                                             \\
            & GAT-like P  & {\cellcolor[rgb]{0.796,0.796,0.796}}91.62 $\pm$ 0.40          & {\cellcolor[rgb]{0.757,0.592,0.153}}\textbf{95.92 $\pm$ 0.46} & {\cellcolor[rgb]{0.859,0.859,0.859}}95.69 $\pm$ 0.29          & {\cellcolor[rgb]{0.898,0.808,0.537}}97.15 $\pm$ 0.13          & {\cellcolor[rgb]{0.592,0.592,0.592}}85.34 $\pm$ 0.68          & {\cellcolor[rgb]{0.784,0.643,0.212}}94.83 $\pm$ 0.18          & 5.50                                               & {\cellcolor[rgb]{0.788,0.788,0.788}}37.45 $\pm$ 1.19         & {\cellcolor[rgb]{0.973,0.918,0.761}}46.12 $\pm$ 3.63          & {\cellcolor[rgb]{0.824,0.824,0.824}}39.98 $\pm$ 2.04          & {\cellcolor[rgb]{0.843,0.718,0.365}}84.94 $\pm$ 0.73          & {\cellcolor[rgb]{0.58,0.58,0.58}}83.20 $\pm$ 0.79             & {\cellcolor[rgb]{0.827,0.698,0.325}}85.32 $\pm$ 0.41         & 8.17                                               & 6.83                                                                      \\ 
            \hline
            \multirow{3}{*}{\begin{tabular}[c]{@{}c@{}}GNNFormer\\PT+PT\end{tabular}} & GCN-like P  & {\cellcolor[rgb]{0.635,0.635,0.635}}92.01 $\pm$ 0.36          & {\cellcolor[rgb]{0.996,0.953,0.831}}95.55 $\pm$ 0.17          & {\cellcolor[rgb]{0.655,0.655,0.655}}\underline{95.79 $\pm$ 0.23}  & {\cellcolor[rgb]{0.769,0.616,0.18}}\underline{97.24 $\pm$ 0.12}  & {\cellcolor[rgb]{0.957,0.957,0.957}}84.73 $\pm$ 0.61          & {\cellcolor[rgb]{0.996,0.953,0.831}}94.54 $\pm$ 0.32          & 8.33                                               & {\cellcolor[rgb]{0.753,0.753,0.753}}37.53 $\pm$ 0.87         & {\cellcolor[rgb]{0.996,0.953,0.831}}45.90 $\pm$ 2.26          & {\cellcolor[rgb]{0.8,0.8,0.8}}40.20 $\pm$ 2.78                & {\cellcolor[rgb]{0.757,0.592,0.153}}\textbf{85.66 $\pm$ 0.69} & {\cellcolor[rgb]{0.749,0.749,0.749}}79.38 $\pm$ 0.43          & {\cellcolor[rgb]{0.78,0.631,0.2}}85.40 $\pm$ 0.35            & 8.33                                               & 8.33                                                                      \\
                        & SAGE-like P & {\cellcolor[rgb]{0.753,0.753,0.753}}91.73 $\pm$ 0.38          & {\cellcolor[rgb]{0.906,0.82,0.561}}95.69 $\pm$ 0.34           & {\cellcolor[rgb]{0.839,0.839,0.839}}95.70 $\pm$ 0.20          & {\cellcolor[rgb]{0.937,0.871,0.663}}97.12 $\pm$ 0.13          & {\cellcolor[rgb]{0.718,0.718,0.718}}85.14 $\pm$ 0.74          & {\cellcolor[rgb]{0.812,0.678,0.29}}94.79 $\pm$ 0.29           & 8.17                                               & {\cellcolor[rgb]{0.545,0.545,0.545}}\textbf{37.99 $\pm$ 1.31} & {\cellcolor[rgb]{0.925,0.847,0.627}}46.52 $\pm$ 3.39          & {\cellcolor[rgb]{0.808,0.808,0.808}}40.13 $\pm$ 2.68          & {\cellcolor[rgb]{0.925,0.843,0.62}}84.22 $\pm$ 0.54           & {\cellcolor[rgb]{0.624,0.624,0.624}}82.21 $\pm$ 0.44          & {\cellcolor[rgb]{0.792,0.651,0.227}}85.38 $\pm$ 0.29         & \textbf{6.17}                                        & 7.17                                                                      \\
                        & GAT-like P  & {\cellcolor[rgb]{0.784,0.784,0.784}}91.65 $\pm$ 0.36          & {\cellcolor[rgb]{0.906,0.82,0.561}}95.69 $\pm$ 0.47           & {\cellcolor[rgb]{0.8,0.8,0.8}}95.72 $\pm$ 0.29                & {\cellcolor[rgb]{0.937,0.871,0.663}}97.13 $\pm$ 0.15          & {\cellcolor[rgb]{0.62,0.62,0.62}}85.30 $\pm$ 0.83             & {\cellcolor[rgb]{0.784,0.643,0.212}}94.83 $\pm$ 0.36          & 7.17                                               & {\cellcolor[rgb]{0.624,0.624,0.624}}37.82 $\pm$ 0.99         & {\cellcolor[rgb]{0.957,0.894,0.718}}46.24 $\pm$ 3.81          & {\cellcolor[rgb]{0.804,0.804,0.804}}40.16 $\pm$ 2.22          & {\cellcolor[rgb]{0.765,0.612,0.173}}85.57 $\pm$ 0.72          & {\cellcolor[rgb]{0.655,0.655,0.655}}81.54 $\pm$ 0.63          & {\cellcolor[rgb]{0.792,0.655,0.243}}85.37 $\pm$ 0.37         & 6.50                                                & 6.83                                                                      \\ 
            \hline
            \multirow{3}{*}{\begin{tabular}[c]{@{}c@{}}GNNFormer\\TT+PP\end{tabular}} & GCN-like P  & {\cellcolor[rgb]{0.678,0.678,0.678}}91.91 $\pm$ 0.49          & {\cellcolor[rgb]{0.792,0.655,0.235}}95.86 $\pm$ 0.28          & {\cellcolor[rgb]{0.545,0.545,0.545}}\textbf{95.84 $\pm$ 0.19} & {\cellcolor[rgb]{0.843,0.722,0.369}}97.19 $\pm$ 0.13          & {\cellcolor[rgb]{0.569,0.569,0.569}}\underline{85.38 $\pm$ 0.77}  & {\cellcolor[rgb]{0.937,0.871,0.663}}94.62 $\pm$ 0.26          & 5.00                                               & {\cellcolor[rgb]{0.847,0.847,0.847}}37.31 $\pm$ 1.00         & {\cellcolor[rgb]{0.867,0.749,0.427}}47.08 $\pm$ 3.92          & {\cellcolor[rgb]{0.631,0.631,0.631}}41.60 $\pm$ 2.30          & {\cellcolor[rgb]{0.878,0.773,0.463}}84.65 $\pm$ 0.66          & {\cellcolor[rgb]{0.847,0.847,0.847}}77.06 $\pm$ 0.77          & {\cellcolor[rgb]{0.835,0.71,0.341}}85.31 $\pm$ 0.28          & 8.67                                                & 6.83                                                                      \\
                        & SAGE-like P & {\cellcolor[rgb]{0.741,0.741,0.741}}91.75 $\pm$ 0.48          & {\cellcolor[rgb]{0.878,0.776,0.486}}95.73 $\pm$ 0.41          & {\cellcolor[rgb]{0.918,0.918,0.918}}95.66 $\pm$ 0.15          & {\cellcolor[rgb]{0.914,0.827,0.584}}97.14 $\pm$ 0.17          & {\cellcolor[rgb]{0.58,0.58,0.58}}85.36 $\pm$ 0.55             & {\cellcolor[rgb]{0.824,0.694,0.31}}94.78 $\pm$ 0.32           & 7.17                                               & {\cellcolor[rgb]{0.663,0.663,0.663}}37.74 $\pm$ 1.00         & {\cellcolor[rgb]{0.757,0.592,0.153}}\textbf{47.98 $\pm$ 4.60} & {\cellcolor[rgb]{0.616,0.616,0.616}}\underline{41.75 $\pm$ 3.06}  & {\cellcolor[rgb]{0.996,0.953,0.831}}83.58 $\pm$ 0.56          & {\cellcolor[rgb]{0.573,0.573,0.573}}83.35 $\pm$ 0.52          & {\cellcolor[rgb]{0.996,0.953,0.831}}85.03 $\pm$ 0.62         & \textbf{6.17}                                       & 6.67                                                                      \\
                        & GAT-like P  & {\cellcolor[rgb]{0.604,0.604,0.604}}\underline{92.08 $\pm$ 0.15}  & {\cellcolor[rgb]{0.78,0.635,0.208}}95.88 $\pm$ 0.43           & {\cellcolor[rgb]{0.859,0.859,0.859}}95.69 $\pm$ 0.25          & {\cellcolor[rgb]{0.929,0.847,0.631}}97.13 $\pm$ 0.20          & {\cellcolor[rgb]{0.608,0.608,0.608}}85.32 $\pm$ 0.47          & {\cellcolor[rgb]{0.812,0.678,0.29}}94.79 $\pm$ 0.28           & 5.00                                               & {\cellcolor[rgb]{0.788,0.788,0.788}}37.45 $\pm$ 0.82         & {\cellcolor[rgb]{0.878,0.773,0.463}}46.97 $\pm$ 4.23          & {\cellcolor[rgb]{0.635,0.635,0.635}}41.57 $\pm$ 2.03          & {\cellcolor[rgb]{0.878,0.773,0.463}}84.65 $\pm$ 0.71          & {\cellcolor[rgb]{0.573,0.573,0.573}}\underline{83.41 $\pm$ 0.51}  & {\cellcolor[rgb]{0.957,0.894,0.714}}85.10 $\pm$ 0.40         & 6.67                                                & \underline{5.83}                                                              \\ 
            \hline
            \multirow{3}{*}{\begin{tabular}[c]{@{}c@{}}GNNFormer\\PP+TT\end{tabular}} & GCN-like P  & {\cellcolor[rgb]{0.659,0.659,0.659}}91.95 $\pm$ 0.41          & {\cellcolor[rgb]{0.914,0.831,0.588}}95.68 $\pm$ 0.16          & {\cellcolor[rgb]{0.878,0.878,0.878}}95.68 $\pm$ 0.16          & {\cellcolor[rgb]{0.929,0.847,0.631}}97.13 $\pm$ 0.14          & {\cellcolor[rgb]{0.78,0.78,0.78}}85.03 $\pm$ 0.78             & {\cellcolor[rgb]{0.925,0.847,0.624}}94.64 $\pm$ 0.40          & 9.17                                               & {\cellcolor[rgb]{0.957,0.957,0.957}}37.05 $\pm$ 0.98         & {\cellcolor[rgb]{0.953,0.886,0.702}}46.29 $\pm$ 2.28          & {\cellcolor[rgb]{0.635,0.635,0.635}}41.57 $\pm$ 3.03          & {\cellcolor[rgb]{0.949,0.878,0.686}}84.01 $\pm$ 0.74          & {\cellcolor[rgb]{0.957,0.957,0.957}}74.48 $\pm$ 1.13          & {\cellcolor[rgb]{0.878,0.776,0.486}}85.23 $\pm$ 0.27         & 12.00                                              & 10.58                                                                     \\
                        & SAGE-like P & {\cellcolor[rgb]{0.639,0.639,0.639}}92.00 $\pm$ 0.34          & {\cellcolor[rgb]{0.835,0.714,0.349}}95.80 $\pm$ 0.48          & {\cellcolor[rgb]{0.898,0.898,0.898}}95.67 $\pm$ 0.17          & {\cellcolor[rgb]{0.996,0.953,0.831}}97.08 $\pm$ 0.14          & {\cellcolor[rgb]{0.6,0.6,0.6}}85.33 $\pm$ 0.59                & {\cellcolor[rgb]{0.855,0.737,0.4}}94.74 $\pm$ 0.33            & 7.83                                               & {\cellcolor[rgb]{0.62,0.62,0.62}}\underline{37.84 $\pm$ 0.85}            & {\cellcolor[rgb]{0.902,0.812,0.545}}46.74 $\pm$ 3.92          & {\cellcolor[rgb]{0.627,0.627,0.627}}41.62 $\pm$ 3.55          & {\cellcolor[rgb]{0.973,0.918,0.765}}83.79 $\pm$ 0.73          & {\cellcolor[rgb]{0.651,0.651,0.651}}81.65 $\pm$ 0.46          & {\cellcolor[rgb]{0.91,0.824,0.576}}85.18 $\pm$ 0.51          & 6.67                                               & 7.25                                                                      \\
                        & GAT-like P  & {\cellcolor[rgb]{0.957,0.957,0.957}}91.22 $\pm$ 0.70          & {\cellcolor[rgb]{0.847,0.729,0.388}}95.78 $\pm$ 0.36          & {\cellcolor[rgb]{0.776,0.776,0.776}}95.73 $\pm$ 0.25          & {\cellcolor[rgb]{0.957,0.89,0.71}}97.11 $\pm$ 0.22            & {\cellcolor[rgb]{0.675,0.675,0.675}}85.21 $\pm$ 0.56          & {\cellcolor[rgb]{0.839,0.714,0.353}}94.76 $\pm$ 0.28          & 9.00                                               & {\cellcolor[rgb]{0.796,0.796,0.796}}37.43 $\pm$ 0.95         & {\cellcolor[rgb]{0.973,0.918,0.761}}46.12 $\pm$ 3.68          & {\cellcolor[rgb]{0.545,0.545,0.545}}\textbf{42.29 $\pm$ 2.50} & {\cellcolor[rgb]{0.812,0.675,0.282}}85.19 $\pm$ 0.74          & {\cellcolor[rgb]{0.627,0.627,0.627}}82.10 $\pm$ 0.82          & {\cellcolor[rgb]{0.871,0.761,0.447}}85.25 $\pm$ 0.47         & 7.00                                               & 8.00                                                                      \\
            \hline                       
    \end{tabular}}
\end{table*}

\subsection{Evaluation on Node Classification}


Table~\ref{tab: main} reports the node classification results of all methods, from which we can draw the following conclusions: 

1) \textbf{Overall performance:} GNNFormer consistently exhibits superior local and global average performance rankings on both homophilous and heterophilous datasets, surpassing the three types of baselines significantly in terms of performance superiority and stability;

2) \textbf{Better universality for transformation-first message passing:}
A more universal GNNFormer architecture tends to prioritize a message passing settings where transformation precedes propagation (i.e., $\mathbf{TPTP}$, $\mathbf{TTPP}$) for node representation learning. This is supported by the higher local and global performance rankings of GNNFormer(TP+TP, TT+PP).
A reasonable explanation is that the transformation operation can map different types of node features into a unified feature space, achieving feature alignment, while also enhancing the expressive power of features through linear transformation and non-linear activation. Therefore, the strategy of transformation followed by propagation improves the model’s adaptability and performance across different types of graphs by unifying and enhancing node/edge feature representations before message propagation, resulting in better universality in both homophilous and heterophilous scenarios.

3) \textbf{More significant preferences in heterophilous scenarios:} Due to the diversity of heterophilous scenarios, the optimal architecture of GNNFormer varies for different heterophilous datasets. For example, on Squirrel-fix, GNNFormer tends to perform multiple propagations ($\mathbf{PP}$) to capture higher-order homophily, or multiple transformations ($\mathbf{TT}$) to adjust and transform features to adapt to heterophily. In contrast, on Tolokers and Penn94, GNNFormer prefers alternating transformation and propagation to adapt to heterophily via immediate feature adjustment and transformation. These observations drive us to explore the automatic search for GNNFormer architectures in future work.

\begin{table*}
    \centering
    \caption{Ablation analysis of the GNNFormer.}
    \label{tab:ablation}
    \renewcommand\arraystretch{1.2}
    \resizebox{\textwidth}{!}{
    \begin{tabular}{c|cccccc|cccccc|c} 
    \hline
    \diagbox{GNNFormer}{Dataset}                                          & Computers                                                 & Photo                                                     & \begin{tabular}[c]{@{}c@{}}Coauthor\\CS\end{tabular}      & \begin{tabular}[c]{@{}c@{}}Coauthor\\Physics\end{tabular} & Wiki-CS                                                    & Facebook                                                   & Actor                                                      & \begin{tabular}[c]{@{}c@{}}Chameleon\\-fix\end{tabular}    & Squirrel-fix                                              & Tolokers                                                  & \begin{tabular}[c]{@{}c@{}}Roman\\-empire\end{tabular}     & Penn94                                                    & \begin{tabular}[c]{@{}c@{}}Global\\Rank\end{tabular}  \\ 
    \hline
    \multirow{2}{*}{Best architecture} & \begin{tabular}[c]{@{}c@{}}TP+TP\\GCN-like P\end{tabular} & \begin{tabular}[c]{@{}c@{}}TP+TP\\GAT-like P\end{tabular} & \begin{tabular}[c]{@{}c@{}}TT+PP\\GCN-like P\end{tabular} & \begin{tabular}[c]{@{}c@{}}TP+TP\\GCN-like P\end{tabular} & \begin{tabular}[c]{@{}c@{}}TP+TP\\SAGE-like P\end{tabular} & \begin{tabular}[c]{@{}c@{}}TP+TP\\SAGE-like P\end{tabular} & \begin{tabular}[c]{@{}c@{}}PT+PT\\SAGE-like P\end{tabular} & \begin{tabular}[c]{@{}c@{}}TT+PP\\SAGE-like P\end{tabular} & \begin{tabular}[c]{@{}c@{}}PP+TT\\GAT-like P\end{tabular} & \begin{tabular}[c]{@{}c@{}}PT+PT\\GCN-like P\end{tabular} & \begin{tabular}[c]{@{}c@{}}TP+TP\\SAGE-like P\end{tabular} & \begin{tabular}[c]{@{}c@{}}TP+TP\\GCN-like P\end{tabular} & \multirow{2}{*}{\textbf{1.17~}}                        \\
                                       & \textbf{92.21 $\pm$ 0.35}                                     & \textbf{95.92 $\pm$ 0.46}                                     & \textbf{95.84 $\pm$ 0.19}                                     & \textbf{97.25 $\pm$ 0.15}                                     & \textbf{85.41 $\pm$ 0.62}                                      & \textbf{94.87 $\pm$ 0.18}                                      & \textbf{37.99 $\pm$ 1.31}                                      & 47.98 $\pm$ 4.60                                               & \textbf{42.29 $\pm$ 2.50}                                     & \textbf{85.66 $\pm$ 0.69}                                     & 83.91 $\pm$ 0.55                                               & \textbf{85.44 $\pm$ 0.36}                                     &                                                        \\ 
    \hline
    w/o FFN                            & 92.05 $\pm$ 0.44                                              & 95.77 $\pm$ 0.28                                              & 95.29 $\pm$ 0.38                                              & 96.98 $\pm$ 0.24                                              & 85.13 $\pm$ 0.53                                               & 94.87 $\pm$ 0.29                                               & 36.57 $\pm$ 0.91                                               & 47.25 $\pm$ 4.63                                               & 41.75 $\pm$ 3.74                                              & 85.54 $\pm$ 0.75                                              & 82.12 $\pm$ 0.32                                               & 85.01 $\pm$ 0.48                                              & 3.92~                                                  \\
    FFN(GEGLU)                         & 92.00 $\pm$ 0.34                                              & 95.81 $\pm$ 0.34                                              & 95.17 $\pm$ 0.37                                              & 97.23 $\pm$ 0.15                                              & 85.17 $\pm$ 0.50                                               & 94.79 $\pm$ 0.24                                               & 37.66 $\pm$ 1.13                                               & 47.53 $\pm$ 4.22                                               & 40.56 $\pm$ 2.11                                              & 85.46 $\pm$ 0.49                                              & 83.69 $\pm$ 0.45                                               & 85.40 $\pm$ 0.36                                              & 3.75~                                                  \\
    FFN(ReGLU)                         & 92.11 $\pm$ 0.43                                              & 95.90 $\pm$ 0.26                                              & 95.15 $\pm$ 0.40                                              & 97.21 $\pm$ 0.14                                              & 85.10 $\pm$ 0.63                                               & 94.75 $\pm$ 0.25                                               & 37.72 $\pm$ 1.05                                               & \textbf{48.26 $\pm$ 2.83}                                      & 41.08 $\pm$ 3.10                                              & 85.60 $\pm$ 0.91                                              & 83.63 $\pm$ 0.54                                               & 85.34 $\pm$ 0.42                                              & 3.33~                                                  \\ 
    \hline
    w/o AIRes                          & 92.06 $\pm$ 0.45                                              & 93.34 $\pm$ 0.65                                              & 95.18 $\pm$ 0.38                                              & 97.25 $\pm$ 0.16                                              & 83.60 $\pm$ 0.49                                               & 94.62 $\pm$ 0.35                                               & 36.02 $\pm$ 1.06                                               & 47.87 $\pm$ 4.27                                               & 41.98 $\pm$ 2.39                                              & 85.52 $\pm$ 0.57                                              & 80.81 $\pm$ 0.88                                               & 85.38 $\pm$ 0.38                                              & 4.25~                                                  \\
    AIRes - Res                        & 91.80 $\pm$ 0.26                                              & 95.35 $\pm$ 0.65                                              & 95.03 $\pm$ 0.54                                              & 96.82 $\pm$ 0.24                                              & 85.08 $\pm$ 0.57                                               & 94.84 $\pm$ 0.20                                               & 36.82 $\pm$ 1.16                                               & 47.98 $\pm$ 3.87                                               & 42.13 $\pm$ 2.61                                              & 79.13 $\pm$ 6.53                                              & \textbf{84.35 $\pm$ 0.56}                                      & 81.87 $\pm$ 0.72                                              & 4.33~                                                  \\
    \hline
    \end{tabular}}
    \end{table*}

\begin{figure}[!htb]
  \centering
  \includegraphics[width=\linewidth]{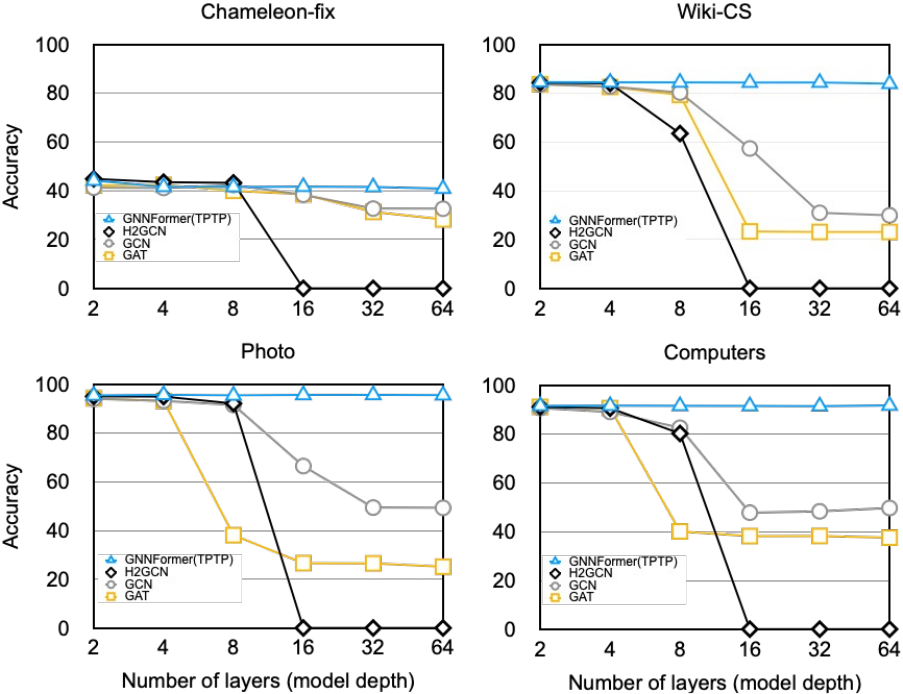}
  \caption{Impact of model depth.}
  \label{fig: over}
\end{figure}

\subsection{More Analysis}

\subsubsection{Impact on FFN}
We further analyze the impact of the FFN module on GNNFormer, as shown in Table~\ref{tab:ablation}. It is evident that when the FFN module is removed (w/o FFN), the model experiences significant performance degradation across all datasets, indicating that the FFN can effectively enhance the expressive power of GNNFormer. Furthermore, we compare the FFN equipped with GEGLU and ReGLU, and observe that FFN with SwishGLU generally maintains higher classification performance in most cases. This is because SwishGLU combines the smoothness of Swish with the flexibility of GLU---the former stabilizes gradient updates, while the latter helps GNNFormer selectively retain and propagate more informative features. Therefore, GNNFormer, utilizing SwishGLU, exhibits better non-linear expressive power, allowing it to capture complex node relationships and feature interactions within various graphs.

\subsubsection{Impact on Residual Connection}

We further conduct an ablation analysis on the Adaptive Initial Residual Connection (AIRes) in the GNNFormer architecture, as shown in Table~\ref{tab:ablation}. When this component is removed (w/o AIRes), GNNFormer experiences a significant performance degradation, indicating that leveraging initial features can substantially enhance GNNFormer's representation learning ability in diverse scenarios. Additionally, when we replace the AIRes with a standard residual connection, the model exhibits even more significant performance degradation in most cases. This suggests that while initial features are important, improperly utilizing them can introduce noise and interfere with the model’s representation learning.

\subsubsection{Impact on Model Depth}

To validate the immunity of our GNNFormer to over-smoothing, we conduct model depth experiments on four datasets. Specifically, we compare GNNFormer with Vanilla GNNs (GCN, GAT) and higher-order GNN method (H2GNN). The latter acquires higher-order information by stacking message-passing modules, whereas our GNNFormer achieves this by stacking PT-blocks while employing AIRes. Figure~\ref{fig: over} shows the performance of all methods at different model depths. It is evident that the performance of Vanilla GNNs rapidly declines as model depth increases, indicating the presence of over-smoothing. For H2GNN, performance gradually decreases as the model depth increases from 2 to 8 layers, and the model experiences memory overflow when the depth exceeds 16 layers. In contrast, our GNNFormer consistently maintains stable performance when stacking PT-block message-passing modules, indicating its immunity to the over-smoothing problem.

\subsubsection{Efficiency Analysis}

\begin{figure}[!htb]
  \centering
  \includegraphics[width=\linewidth]{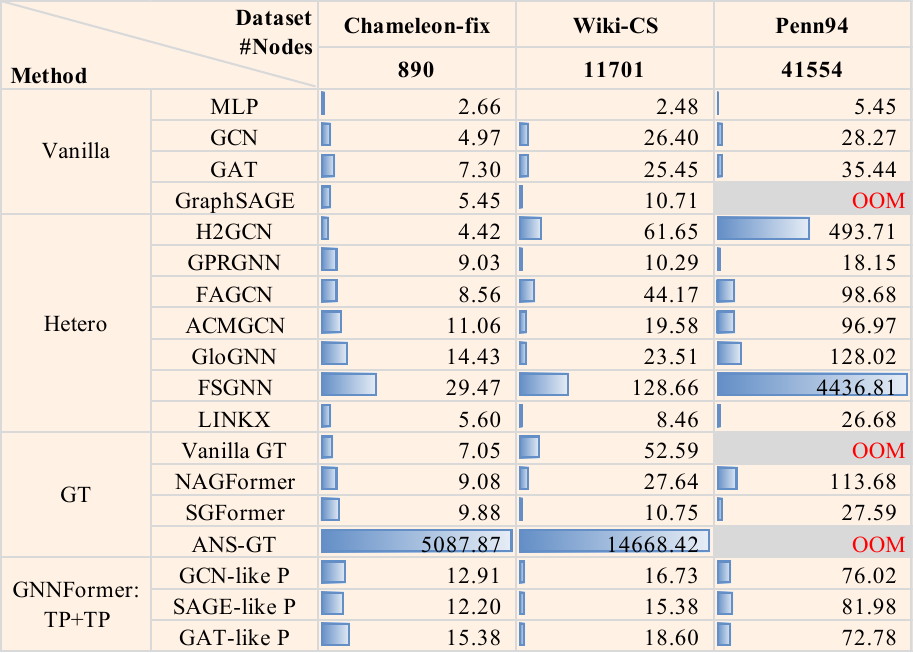}
  \caption{Average running time per epoch (ms).}
  \label{fig: time}
\end{figure}

The efficiency analysis of all baselines and GNNFormer (TPTP) is shown in Figure~\ref{fig: time}. In the small-scale dataset Chameleon-fix, the runtime advantage of GNNFormer is not particularly evident. In Wiki-CS, where the number of nodes exceeds 10,000, the average runtime of GT methods increases by more than 2.5 times compared to Chameleon-fix, while the runtime of our GNNFormer remains relatively stable. In the Penn94 dataset, which contains over 40,000 nodes, GT methods involving quadratic complexity in attention computation encounters OOM issues, whereas the runtime of GNNFormer remains at a relatively low level. Therefore, our GNNFormer architecture demonstrates a clear advantage in computational efficiency compared to traditional GT architectures.

\section{Conclusion}

To demonstrate the inherent misalignment of Graph Transformers in node classification tasks, we start with observational experiments and progressively develop the GNNFormer framework. Extensive experiments on 12 benchmark datasets demonstrate that, compared to existing Graph Transformer methods, our framework not only maintains superior performance but also improves computational efficiency. Additionally, we find that GNNFormer, which employs a transformation-first message-passing manner exhibits greater universality in both homophilous and heterophilous scenarios. This insight also motivates us to explore the automatic search of GNNFormer architectures in future work, i.e., driving GNNFormer to automatically search for the optimal combination of transformation and propagation operations, which will further enhance the generalizability and superiority of our framework.

\section*{Acknowledgement}
This work was supported in part by the Key R\&D Program of Zhejiang under Grants 2022C01018 and 2024C01025, by the National Natural Science Foundation of China under Grants 62103374 and U21B2001.

\bibliography{aaai25}


\newpage
~
\newpage
\appendix
\section*{Appendix}
\subsection{A. Dataset Details}\label{app: dataset}
\begin{itemize}
    \item Computers and Photo~\cite{mcauley2015I} are segments of the Amazon co-purchase graph, where nodes represent products, edges represent the co-purchased relations of products, and features are bag-of-words vectors extracted from product reviews.
    \item Coauthor CS and Coauthor Physics~\cite{Shchur2018PitfallsOG} are co-authorship graphs based on the Microsoft Academic Graph from the KDD Cup 2016 challenge, where nodes represent authors, edge represent the corresponding authors have co-authored a paper, features consist of keywords from each author's published papers, and the class labels denote the most active research fields for each author.
    \item Wiki-CS~\cite{mernyei2020wiki} is a Wikipedia-based dataset which is constructed from Wikipedia categories, specifically 10 classes corresponding to branches of computer science, with very high connectivity. The node features are derived from the text of the corresponding articles. They were calculated as the average of pretrained GloVe word embeddings, resulting in 300-dimensional node features.
    \item Facebook~\cite{rozemberczki20201R} is a page-page graph of verified Facebook sites, where nodes correspond to official Facebook pages, links to mutual likes between sites, and features are extracted from the site descriptions. 
    \item Actor~\cite{tang2009S} is a network dataset designed for analyzing co-occurrence relationships among actors, where node represents an actor, and the edges between nodes indicate their co-occurrence on the same Wikipedia page. The node features are constructed from keywords extracted from the respective actors' Wikipedia pages.
    \item Chameleon-fix and Squirrel-fix~\cite{platonov2023a} are two page-page networks focusing on specific topics in Wikipedia, where nodes represent web pages, and edges denote mutual links between the pages. The node features are composed of informative nouns extracted from the corresponding Wikipedia pages. The task of these datasets is to categorize the nodes into five distinct groups based on the average monthly traffic received by each web page.
    \item Tolokers~\cite{platonov2023a} is a social network extracted from the Toloka crowdsourcing platform, where nodes represent workers and two workers are connected if they participate in the same task. The node features are constructed from the workers' profile information and task performance statistics, while the labels indicate whether a worker is banned in a project.
    \item Roman-empire~\cite{platonov2023a} is derived from the Roman Empire article on Wikipedia, where nodes in the dataset represent words from the article, edges indicating word dependencies. The node features are constructed from word embeddings obtained using the FastText method, and labels denote the syntactic roles of the words.
    \item Penn94~\cite{NEURIPS2021_ae816a80} is a Facebook social network, where nodes denote students and are labeled with the gender of users, edges represent the relationship of students. Node features
    are constructed from basic information about students which are major, second major/minor, dorm/house, year and high school.
\end{itemize}
\begin{table}[!htb]
\renewcommand\arraystretch{1.3}
\centering
\caption{Summary of datasets used}
\label{tab: datasets}
\resizebox{\linewidth}{!}{
    \begin{tabular}{c|cccc} 
\hline
                 & Node Feature & Node Number & Edges   & Classes  \\ 
\hline
Computers        & 767          & 13752       & 491722  & 10       \\
Photo            & 745          & 7650        & 238162  & 8        \\
Coauthor CS      & 6805         & 18333       & 163788  & 15       \\
Coauthor Physics & 8415         & 34493       & 495924  & 5        \\
Wiki-CS          & 300          & 11701       & 431726  & 10       \\
Facebook         & 128          & 22470       & 342004  & 4        \\
Actor            & 932          & 7600        & 30019   & 5        \\
Chameleon-fix    & 2325         & 890         & 13584   & 5        \\
Squirrel-fix     & 2089         & 2223        & 65718   & 5        \\
Tolokers         & 10           & 11758       & 1038000 & 2        \\
Roman-empire     & 300          & 22662       & 65854   & 18       \\
Penn94           & 4814         & 41554       & 2724458 & 2        \\
\hline
\end{tabular}
}
\end{table}
\subsection{B. Baseline Details}\label{app: baselines}
\begin{table*}[!htb]
  \renewcommand\arraystretch{1.3}
  \centering
  \caption{Optimal parameters for GNNFormer}
  \label{tab:bestparameters}
  \resizebox{\textwidth}{!}{
      \begin{tabular}{c|cccccccccccc} 
\hline
\multicolumn{1}{l|}{} & Computers  & Photo      & Coauthor CS & Coauthor Physics & Wiki-CS     & Facebook    & Actor       & Chameleon-fix & Squirrel-fix & Tolokers   & Roman-empire & Penn94      \\ 
\hline
P-like                & GCN-like P & GAT-like P & GCN-like P  & GCN-like P       & SAGE-like P & SAGE-like P & SAGE-like P & SAGE-like P   & GAT-like P   & GCN-like P & SAGE-like P  & GCN-like P  \\
model type            & TPTP       & TPTP       & TTPP        & TPTP             & TPTP        & TPTP        & PTPT        & TTPP          & PPTT         & PTPT       & TPTP         & TPTP        \\
learning rate         & 5e-3       & 5e-3       & 1e-2        & 5e-3             & 5e-3        & 5e-3        & 5e-3        & 5e-3          & 1e-2         & 5e-2       & 5e-3         & 5e-3        \\
weight decay          & 5e-4       & 1e-5       & 1e-4        & 5e-5             & 5e-4        & 1e-4        & 1e-4        & 5e-4          & 5e-4         & 5e-3       & 5e-4         & 5e-5        \\
dropout               & 0.7        & 0.5        & 0.1         & 0.7              & 0.3         & 0.5         & 0.5         & 0.9           & 0.5          & 0.3        & 0.3          & 0.5         \\
\hline
\end{tabular}
  }
\end{table*}
\begin{itemize}
    \item 
    $\textbf{MLP}$ is a two-layer linear neural network that based on the original features of the nodes, without any propagation or aggregation rules.
    \item 
    \textbf{GCN} is a neural network that aggregates information among neighboring nodes through message passing.
    \item 
    \textbf{GAT} is a neural network that leverages multi-head attention to weight node features effectively on graph data.
    \item 
    \textbf{SAGE} is a graph neural network that learns node representations by sampling and aggregating neighborhood information.
    \item \textbf{H2GCN} constructs a neural network by separating ego and neighbor embeddings, aggregating higher-order neighborhood information, and combing intermediate representations.
    \item \textbf{GPRGNN} is a graph neural network that optimizes node feature and topology extraction by adaptively learning Generalized PageRank weights.
    \item \textbf{FAGCN} is a novel graph convolutional network that integrates low and high-frequency signals through an adaptive gating mechanism.
    \item \textbf{ACMGCN} adaptively employs aggregation, diversification, and identity channels to extract richer local information for each node at every layer.
    \item \textbf{GloGNN} generates node embeddings by aggregating global node information and effectively captures homophily by learning a correlation matrix between nodes.
    \item \textbf{FSGNN} is a simplified graph neural network model that enhances node classification performance by introducing a soft selection mechanism.
    \item \textbf{LINKX} combines independent embeddings of the adjacency matrix and node features, generating predictions through a multi-layer perceptron and simple transformations.
    \item \textbf{ANS-GT} is a graph transformer architecture that effectively captures long-range dependencies and global context information through adaptive node sampling and hierarchical graph attention mechanisms.
    \item \textbf{NAGFormer} is a novel graph transformer that handles node classification tasks on large graphs by treating each node as a sequence aggregated from features of neighbors at various hops.
    \item \textbf{SGFormer} is a simplified and efficient graph transformer model that handles large-scale graph data through a single-layer global attention mechanism, achieving node representation learning with linear complexity.
\end{itemize}
\subsection{C. More Parameter Settings}\label{app: parameter}
    We used the Neural Network Intelligence (NNI) tool for hyper-parameter tuning to conduct experiments on the baseline models. The experiments were conducted using the same base parameters as our method, along with specific parameters unique to each baseline model. The special parameters are as follows:
    \begin{itemize}
        \item GloGNN: norm\_layers within \{1,2,3\}, orders within \{2,3,4\}, term weight within \{0,1\}, weighting factor within \{0,1,10\} and \{0.1,1,10,100,1000\} and the balanced term parameters.
        \item FSGNN: aggregator within \{cat, sum\}.
        \item ANS-GT: data\_augmentation within \{4,8,16,32\}, n\_layer within \{2,3,4\} and batch size within \{8,16,32\}.
        \item NAGFormer: hidden within \{128,256,512\}, number of Transformer layers within \{1,2,3,4,5\} and number of propagation steps K \{7,10\}.
        \item SGFormer: number of global attention layers is fixed as 1, number of GCN layers within \{1,2,3\}, weight $\alpha$ within \{0.5,0.8\}.
    \end{itemize}
    Table~\ref{tab:bestparameters} presents the optimal parameters for the GNNFormer framework, encompassing the learning rate, weight decay, and dropout as the key hyperparameters. Additionally, it highlights the optimal model type and the $\mathbf{P}$ operation type.

\end{document}